\documentclass[10pt,twocolumn,letterpaper]{article}

\usepackage[pagenumbers]{cvpr} % To force page numbers, e.g. for an arXiv version

\usepackage{booktabs}  % For \toprule, \midrule, \bottomrule
\usepackage{multirow}  % For \multirow
\usepackage{graphicx}  % For \resizebox
\usepackage{tabularx}

\definecolor{cvprblue}{rgb}{0.21,0.49,0.74}
\usepackage[breaklinks,colorlinks,allcolors=cvprblue]{hyperref}
\usepackage[accsupp]{axessibility}
\def\paperID{41393} % *** Enter the Paper ID here
\def\confName{CVPR}
\def\confYear{2026}

\title{TopoCL: Topological Contrastive Learning for Medical Imaging}

\author{
Guangyu Meng$^{1}$ \qquad
Pengfei Gu$^{2}$\thanks{Corresponding authors. This research was supported in part by NSF Grants CCF-2444309 and CCF-2523787, AHA award 26AIREA1574568.} \qquad
Peixian Liang$^{1}$ \\[0.3em]
John P. Lalor$^{1,3}$ \qquad
Erin Wolf Chambers$^{1}$\footnotemark[1] \qquad
Danny Z. Chen$^{1}$\footnotemark[1]\\[0.5em]
$^{1}$Dept.\ of Computer Science and Engineering, University of Notre Dame\\
$^{2}$The University of Texas Rio Grande Valley \\ \quad
$^{3}$Dept.\ of IT, Analytics, and Operations, University of Notre Dame\\[0.3em]
{\tt\small \{gmeng, pliang, john.lalor, echambe2, dchen\}@nd.edu, \quad pengfei.gu01@utrgv.edu}
}
\begin{document}
\maketitle
\renewcommand{\thefootnote}{\fnsymbol{footnote}}

\begin{abstract}
Contrastive learning (CL) has become a powerful approach for learning representations from unlabeled images. However, existing CL methods focus predominantly on visual appearance features while neglecting topological characteristics (e.g., connectivity patterns, boundary configurations, cavity formations) that provide valuable cues for medical image analysis. To address this limitation, we propose a new topological CL framework (TopoCL) that explicitly exploits topological structures during contrastive learning for medical imaging. Specifically, we first introduce topology-aware augmentations that control topological perturbations using a relative bottleneck distance between persistence diagrams, preserving medically relevant topological properties while enabling controlled structural variations. We then design a Hierarchical Topology Encoder that captures topological features through self-attention and cross-attention mechanisms. Finally, we develop an adaptive mixture-of-experts (MoE) module to dynamically integrate visual and topological representations. TopoCL can be seamlessly integrated with existing CL methods. We evaluate TopoCL on five representative CL methods (SimCLR, MoCo-v3, BYOL, DINO, and Barlow Twins) and five diverse medical image classification datasets. The experimental results show that TopoCL achieves consistent improvements: an average gain of +3.26\% in linear probe classification accuracy with strong statistical significance, verifying its effectiveness. 
Code is publicly available.
\end{abstract}
\vspace{-0.4cm}    
\section{Introduction}
\label{sec:intro}

\begin{figure}[t]
    \centering
    \includegraphics[width=0.82\linewidth,trim={0cm 0.4cm 0.6cm 0cm},clip]{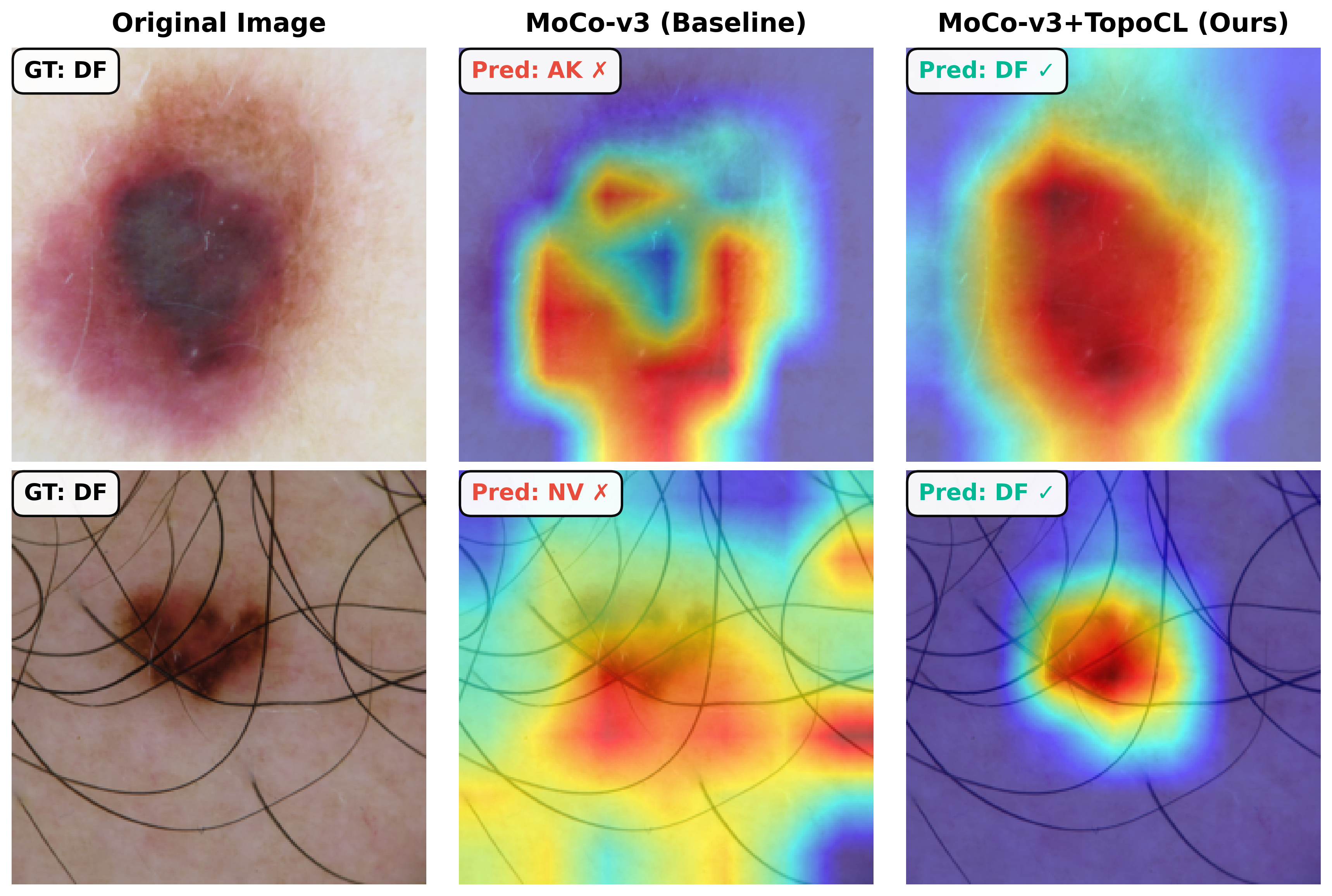}
    \caption{Visualization of failure cases corrected by TopoCL on ISIC2019~\cite{tschandl2018ham10000}. 
    Two dermatofibroma (DF) cases are misclassified by the baseline MoCo-v3~\cite{chen2021mocov3} 
    as actinic keratosis (AK, top) and melanocytic nevi (NV, bottom). MoCo-v3+TopoCL correctly 
    classifies both as DF by capturing characteristic topological features: circular-to-oval 
    boundary patterns with uniform internal connectivity (top) and radial pigmentation structures 
    with a consistent boundary configuration despite hair overlay (bottom). Grad-CAM~\cite{selvaraju2017grad} 
    heatmaps show that TopoCL focuses on lesion boundaries and internal structural patterns, 
    while the baseline exhibits scattered attention on peripheral or irrelevant regions.}
    \label{fig:motivation}
    \vspace{-0.4cm}
\end{figure}

Medical image annotation is a resource-intensive process that requires specialized domain expertise, making it significantly more costly and time-consuming than labeling natural images~\cite{litjens2017survey,irvin2019chexpert,rajpurkar2017chexnet,campanella2019clinical}. As a result, medical image datasets often suffer from severe label scarcity and substantial inter-observer variability~\cite{irvin2019chexpert,rajpurkar2017chexnet}. These labeling challenges fundamentally constrain supervised learning approaches, limiting model generalization and clinical applicability~\cite{zhou2024comprehensive,Azizi_2021_ICCV}. Contrastive learning (CL) has emerged as a powerful approach to address these annotation limitations, enabling models to learn rich visual representations from abundant unlabeled data before fine-tuning for downstream tasks with minimal labeled samples~\cite{shurrab2022self,chen2020simple,he2020momentum,caron2021emerging}.

The core principle of CL is to apply data augmentations to create multiple views of each image, encode these views through deep neural networks, and learn representations that maximize agreement between different views of the same image~\cite{chen2020simple,he2020momentum}. Representative CL methods include SimCLR~\cite{chen2020simple}, MoCo-v3~\cite{chen2021mocov3}, BYOL~\cite{grill2020bootstrap}, DINO~\cite{caron2021emerging}, and Barlow Twins~\cite{zbontar2021barlow}, which employ learning objectives based on contrastive losses, momentum encoders, predictive learning, self-distillation, and redundancy reduction. These methods achieve impressive performance by capturing local appearance features such as textures, intensities, and color patterns. However, they fundamentally operate on pixel-level semantics and do not explicitly encode topological patterns (e.g., connectivity, holes, and boundary configurations), thus neglecting 
%critical 
structural information useful for medical image analysis. As Fig.~\ref{fig:motivation} shows, this limitation can lead to misclassifications of lesion types that differ primarily in boundary patterns, connectivity structures, or cavity formations but are ``similar'' in visual appearance. Such topological differences, captured by persistent homology, may be diagnostically critical but are largely overlooked by visual-only contrastive learning methods.
%causing MoCo-v3 to misclassify both dermatofibroma cases.

To address this gap, we propose \textbf{TopoCL}, a new general topological contrastive learning framework for medical images that augments standard contrastive learning with explicit topology preservation. Since standard CL augmentations provide no guarantee on structural preservation, we first introduce \textit{topology-aware augmentations} that quantify and control topological perturbations using relative bottleneck distance computed on regions of interest. We then design a \textit{Hierarchical Topology Encoder} that processes topological features across different dimensions ($H_0$: connected components; $H_1$: holes) through self-attention and cross-attention mechanisms to capture inter-dimensional structural relationships. Both the visual and topological encoders are pretrained independently using identical contrastive objectives. Finally, we employ an \textit{adaptive Mixture-of-Experts (MoE) fusion module} to dynamically integrate their representations through five complementary experts (visual-only, topology-only, concatenation, gated blending, and cross-attention) using learned per-sample gating, producing the final topology-aware representations.
%ensuring broad compatibility with existing CL methods.

Our main contributions are as follows: {\textbf{(1) Topology-aware augmentation design:} Our new systematic method for designing augmentations explicitly quantifies and controls topological perturbations. Using relative bottleneck distance on regions of interest, we produce weak and strong topological augmentations that preserve diagnostically relevant structures while providing sufficient topological diversity for contrastive learning.}
\textbf{(2) TopoCL framework:} We propose a new topology-enhanced CL framework with a Hierarchical Topology Encoder and an adaptive MoE fusion module for integrating topological priors into contrastive learning.
\textbf{(3) Comprehensive validation:} We demonstrate consistent average accuracy improvements of +3.26\% across five CL methods and five medical image classification benchmarks, with strong statistical significance (86\% of comparisons with $p<0.05$ and 80\% with $p<0.001$). Our code is publicly available at \url{https://github.com/gm3g11/TopoCL}.

\begin{figure*}[t]
    \centering
\includegraphics[width=0.82\textwidth,trim={0cm 0.4cm 0.5cm 0cm},clip]{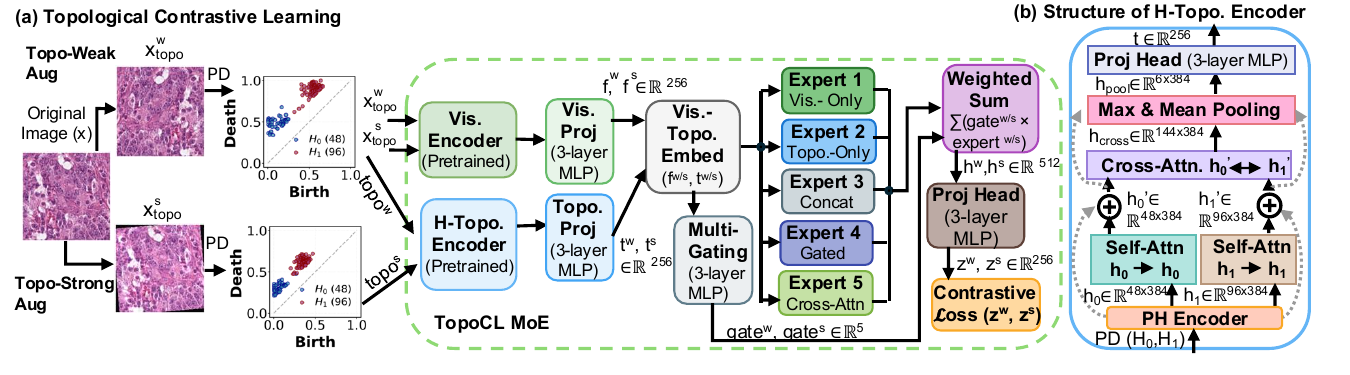}

\caption{An overview of our TopoCL framework. (a) \textbf{Topological Contrastive Learning}: Given an input image $x$, we generate two views with topology-weak ($x_{\text{topo}}^w$) and topology-strong ($x_{\text{topo}}^s$) augmentations, and compute their persistence diagrams $\text{topo}^w$ and $\text{topo}^s$. The visual encoder processes $x_{\text{topo}}^w$, $x_{\text{topo}}^s$ while the topology encoder processes $\text{topo}^w$, $\text{topo}^s$, with projection heads producing visual features $\mathbf{f}^w$, $\mathbf{f}^s$ and topological features $\mathbf{t}^w$, $\mathbf{t}^s$. The TopoCL MoE module fuses these features through five 
%specialized 
experts (the colors distinguish the fusion strategies: Vis-Only, Topo-Only, Concat, Gated, and Cross-Attn) with learned gating weights, producing fused representations $\mathbf{h}^w$, $\mathbf{h}^s$. A projection head maps these to the final representations $\mathbf{z}^w$, $\mathbf{z}^s$, optimized via contrastive loss. (b) \textbf{The structure of H-Topo.~Encoder}: It uses hierarchical self- and cross-attention to capture topological features from PDs with both $H_0$ and $H_1$ homology dimensions.}
\label{fig:overview}
\vspace{-0.4cm}
\end{figure*}

\section{Related Work}
\label{sec:related}

\noindent\textbf{Contrastive Learning in Medical Imaging.} CL has become a powerful paradigm for learning visual representations from unlabeled images by maximizing the agreement between augmented views of the same instance~\cite{shurrab2022self}. Representative CL methods include SimCLR~\cite{chen2020simple} with InfoNCE loss, MoCo-v3~\cite{chen2021mocov3} with momentum encoders, BYOL~\cite{grill2020bootstrap} with online-target asymmetry, Barlow Twins~\cite{zbontar2021barlow} with redundancy reduction, and DINO~\cite{caron2021emerging} with self-distillation. Despite their different loss formulations, these methods share the same fundamental principle of learning by comparing multiple views of the same input, and have been successfully applied to medical imaging~\cite{Azizi_2021_ICCV,tu2024towards}.

However, existing CL methods fundamentally learn visual appearance features from local pixel neighborhoods, overlooking global structural properties critical for medical image analysis. Their augmentation strategies (random cropping, color jittering, Gaussian blurring, etc.) were designed for visual appearances and do not characterize topological structures. Our work addresses these limitations by augmenting CL with explicit topological feature learning and topology-aware augmentations, providing a general enhancement scheme applicable to any CL method.

\noindent\textbf{Topological Data Analysis for Medical Images.}
Appearance-driven CL methods can overlook geometric properties useful for medical image analysis~\cite{peng2024phg,gu2026integrating,gu2025topoimages}, such as tissue components, glandular lumens, and lesion boundaries~\cite{elston1991pathological,lawson2019persistent}. Persistent homology (PH)~\cite{edelsbrunner2010computational} quantifies topology by tracking the births and deaths of connected components ($H_0$) and loops ($H_1$), summarizing them in persistence diagrams (PDs) with stability guarantees~\cite{cohen2005stability,chambers2025stable}. A topological feature in an image $I$ based on a filtration function is characterized by its birth time $b$ and death time $d$, forming $\text{PD}(I) = \{(b_i, d_i)\}$. The bottleneck distance $d_B(\text{PD}_1, \text{PD}_2) = \inf_{\gamma} \sup_{p} \|p - \gamma(p)\|_\infty$ measures dissimilarity between two PDs via an optimal matching $\gamma$~\cite{cohen2005stability}. We employ cubical PH~\cite{wagner2011efficient}, extracting topological signatures from the pixel intensities of images.

Previous work applied PH to medical imaging for texture analysis~\cite{singh2023topological}, segmentation~\cite{clough2020topological,gu2025self,adame2025topo}, tumor grading~\cite{qaiser2019fast,lawson2019persistent}, and supervised classification with architectures like PersLay~\cite{carriere2020perslay} and PHG-Net~\cite{peng2024phg}. These methods predominantly use topology as auxiliary losses or in labeled settings. Unlike these methods, our work integrates PH into contrastive self-supervised learning, introducing a Hierarchical Topology Encoder with cross-attention to capture relationships between $H_0$ and $H_1$ and designing topology-aware augmentations that control structural perturbations.

\noindent\textbf{Feature Fusion Strategies.}
Effective integration of heterogeneous features is critical for representation learning~\cite{zhang2020multimodal,liang2024foundations}. Common strategies include concatenation~\cite{ngiam2011multimodal}, bilinear pooling~\cite{gao2016compact}, cross-attention~\cite{vaswani2017attention}, gated fusion~\cite{arevalo2017gated}, and Transformer-based fusion~\cite{lu2019vilbert}. However, these fixed strategies assume that one approach is optimal for all inputs.

Mixture-of-Experts (MoE) methods~\cite{shazeer2017outrageously,fedus2022switch} provide a framework for adaptive computation by routing inputs to specialized sub-networks. While successful in language models and Vision Transformers~\cite{riquelme2021scaling}, to the best of our knowledge, MoE has not been explored for fusing visual and topological features. Our work introduces the first MoE architecture for visual-topological feature fusion.
\section{Method}
\label{sec:method}

Fig.~\ref{fig:overview} illustrates our TopoCL framework. Our approach enhances standard CL by incorporating topological features: We independently pretrain the visual and topology encoders, and then jointly fine-tune them with adaptive MoE fusion. Specifically, we present an overview of the framework in Section~\ref{sec:overview}, the topology-aware augmentations in Section~\ref{sec:topo_aug}, hierarchical topology encoder in Section~\ref{sec:htopo}, and MoE fusion module in Section~\ref{sec:moe}.

\subsection{Overview}
\label{sec:overview}
Given unlabeled medical images, $\mathcal{D} = \{x_i\}_{i=1}^N$, and a base CL method $\mathcal{M}$ (e.g., SimCLR, MoCo-v3, BYOL), TopoCL enhances $\mathcal{M}$ by incorporating topological features to learn effective representations. We adopt a pretraining-then-fusion strategy: We independently pretrain visual encoder $f_v$ and topology encoder $f_t$, both using contrastive objective $\mathcal{L}_{\mathcal{M}}$ but with different augmentations. For visual pretraining, we apply standard augmentations (random crop, color jitter, Gaussian blur) to generate $x_{\text{vis}}^w$ and $x_{\text{vis}}^s$ (superscripts $w$ and $s$ indicate weak and strong views). For topology pretraining and joint fine-tuning, we apply topology-aware augmentations (Section~\ref{sec:topo_aug}) to generate $x_{\text{topo}}^w$ and $x_{\text{topo}}^s$. Independent pretraining ensures that each encoder learns feature-specific representations without interference, while joint fine-tuning aligns feature spaces for fusion.

\textbf{Visual Encoder Pretraining.} We generate $x_{\text{vis}}^w$ and $x_{\text{vis}}^s$, and optimize the encoder parameters $\theta_v$ through a 3-layer MLP projection head $g_v$:
\begin{equation*}
\theta_v^* = \arg\min_{\theta_v} \mathbb{E}_{x \sim \mathcal{D}} [\mathcal{L}_{\mathcal{M}}(g_v(f_v(x_{\text{vis}}^w)), g_v(f_v(x_{\text{vis}}^s)))].
\end{equation*}

\textbf{Topology Encoder Pretraining.} We generate $x_{\text{topo}}^w$ and $x_{\text{topo}}^s$ and compute their PDs via persistent homology: $\text{topo}^w = \text{PD}(x_{\text{topo}}^w)$ and $\text{topo}^s = \text{PD}(x_{\text{topo}}^s)$. We optimize the encoder parameters $\theta_t$ through a 3-layer MLP projection head $g_t$:
\begin{equation*}
\label{eq:topo_pretrain}
\theta_t^* = \arg\min_{\theta_t} \mathbb{E}_{x \sim \mathcal{D}} [\mathcal{L}_{\mathcal{M}}(g_t(f_t(\text{topo}^w)), g_t(f_t(\text{topo}^s)))].
\end{equation*}

\textbf{Joint Fine-tuning.} After pretraining, we integrate both encoders via an adaptive MoE module (TopoCL MoE in Fig.~\ref{fig:overview}(a)) and jointly fine-tune all parameters, including the pretrained encoders $f_v$ and $f_t$. 
We project both encoder outputs to a shared 256-dimensional embedding space 
through separate 3-layer MLP projection heads, yielding visual features 
$\mathbf{f}^{w/s}$ and topological features $\mathbf{t}^{w/s}$. These features are then fused through the MoE module and optimized using the contrastive loss $\mathcal{L}_{\mathcal{M}}$.

\subsection{Topology-Aware Augmentations}
\label{sec:topo_aug}

Existing CL methods~\cite{chen2020simple,chen2021mocov3,grill2020bootstrap} use standard visual augmentation strategies designed to preserve visual appearances but do not explicitly control their effects on topological structures. However, topological features such as lesion boundaries and tissue connectivity are critical for medical image analysis~\cite{qaiser2019fast,lawson2019persistent,11202494}, and visual-only augmentations can inadvertently alter these medically-relevant structures. To address this lack of topological control, we introduce topology-weak and topology-strong augmentations, which maintain structural similarity for positive pairs while providing sufficient topological diversity for CL methods.

Designing such topology-aware augmentations involves three challenges. (1) We need to define a quantitative measure for topological changes after augmentations. (2) We need to identify which augmentation operations and their parameter settings produce controllable topological changes. (3) We need to establish optimal 
%$d_B^{\text{rel}}$ 
measure ranges for topology-weak versus topology-strong augmentations that balance topological similarity with sufficient diversity for effective TopoCL training.

For the first challenge, we define a relative bottleneck distance to quantify topological changes:
\begin{equation*}
d_B^{\text{rel}}(\mathcal{A}, x) = \frac{d_B(\text{PD}(x), \text{PD}(\mathcal{A}(x)))}{\text{span}(\text{PD}(x))},
\end{equation*}
where $\text{PD}(x)$ denotes the persistence diagram of image $x$, $\mathcal{A}$ is an augmentation operator, $\mathcal{A}(x)$ is the augmented version of image $x$, $d_B$ is the bottleneck distance between PDs, and $\text{span}(\text{PD}(x)) = \max_{(b,d) \in \text{PD}(x)} |d - b|$ provides normalization to enable comparison across images with different topological scales, where $(b,d)$ denotes the birth and death times of topological features. We compute $d_B^{\text{rel}}$ separately for the $H_0$ and $H_1$ homology dimensions, and take the maximum as the overall topological change. Our $d_B^{\text{rel}}$ measure is grounded in the stability theorem~\cite{cohen2005stability}, which guarantees that augmentations with bounded intensity produce bounded changes in PDs.

To apply this measure effectively, we consider where to compute PDs. Computing $d_B^{\text{rel}}$ on full medical images can include background noise and artifacts that obscure 
%diagnostically 
relevant topological changes. Thus, we compute PDs on regions of interest (ROIs) rather than on the full images. We extract ROIs using the Segment Anything Model (SAM)~\cite{kirillov2023segment}, which automatically identifies foreground regions containing relevant structures. Consistent with previous observations~\cite{qaiser2019fast,clough2020topological}, we find that persistent topological features concentrate within ROIs. This ROI-focused approach ensures that $d_B^{\text{rel}}$ captures diagnostically relevant structural changes while filtering out background artifacts.

\begin{table}[t]
\centering
\caption{Topology-aware augmentation categories. All topological effects (except homeomorphism) are parameter-dependent.}
\label{tab:aug_categories}
\scriptsize
\begin{tabularx}{\columnwidth}{p{2cm}p{2cm}X}
\toprule
\textbf{Category} & \textbf{Operations} & \textbf{Topological Effect} \\
\midrule
Homeomorphism & Flips, rotations & Topology-preserving ($d_B^{\text{rel}} \!\approx\! 0$) \\
Boundary perturb. & Gaussian noise & Perturb boundary structures \\
Smoothing & Gaussian blur & Merge/separate features \\
Intensity & Contrast, brightness & Shift feature prominence \\
Morphological & Dilation, erosion & Modify connectivity and holes \\
\bottomrule

\end{tabularx}
\vspace{-0.4cm}
\end{table}

For the second challenge, we identify which operations and parameter settings produce controllable topological changes. However, exhaustive testing across all possible operations is infeasible. Thus, we categorize operations based on their mechanisms and topological effects, and determine representative operations within each category. We identify five categories (Table~\ref{tab:aug_categories}): homeomorphisms (flips, rotations) preserve topology by reorienting pixel grids; boundary perturbations (Gaussian noise) alter sublevel set boundaries, potentially merging or splitting components; smoothing (Gaussian blur) merges nearby features or eliminates small structures; intensity transforms (contrast, brightness) shift filtration thresholds; morphological operations (dilation, erosion) directly modify pixel connectivity, adding or removing connections.

We systematically vary each operation's intensity and measure $d_B^{\text{rel}}$ across configurations, confirming positive correlation between intensity and $d_B^{\text{rel}}$. For example, increasing Gaussian noise can cause boundaries to fragment or merge, substantially increasing $d_B^{\text{rel}}$. We extend this to 2-3 operation combinations, identifying intensity ranges with varying topological changes.

For the third challenge, having determined how operations affect topology with $d_B^{\text{rel}}$, we identify optimal intensity ranges for topology-weak vs topology-strong augmentations. We consider various weak-strong range pairs by training TopoCL and measuring downstream linear probe performance (e.g., weak: 0-10\%, strong: 10-20\%; weak: 5-15\%, strong: 15-25\%). With systematic evaluations, we find that topology-weak augmentations should use operation intensities producing $d_B^{\text{rel}}$ of 5-15\%, and topology-strong augmentations should use intensities giving $d_B^{\text{rel}}$ of 15-25\%. This yields optimal performance across the datasets. Complete ablation results are given in Supplementary Material.

During training, we randomly select an operation combination and sample each operation's intensity from validated ranges corresponding to the desired augmentation type: topology-weak augmentations use ranges producing $d_B^{\text{rel}}$ of 5-15\%, while topology-strong augmentations use 15-25\%. For example, on ISIC2019, applying Flip with Gaussian noise $\sigma \in [0.05, 0.15]$ and contrast in [0.1, 0.2] produces topology-weak augmentations. When this combination is selected, we sample noise and contrast from these ranges without computing $d_B^{\text{rel}}$ at training time.

\subsection{Hierarchical Topology Encoder}
\label{sec:htopo}

Having designed topology-aware augmentations, we consider how to encode PDs into learnable representations for contrastive learning. PDs are unordered sets of birth-death pairs, presenting a permutation invariance challenge similar to point cloud processing~\cite{qi2017pointnet}. Moreover, persistent homology (PH) produces two types of features with distinct geometric semantics: $H_0$ captures connected components while $H_1$ captures holes; $H_0$ and $H_1$ exhibit geometric dependencies (e.g., holes are always bounded by connected components). We design a hierarchical topology encoder (H-Topo.~Encoder) that addresses permutation invariance while respecting such structural distinctions and geometric relationships. Fig.~\ref{fig:overview}(b) shows our H-Topo.~Encoder architecture. Given a PD containing $H_0$ and $H_1$ features, we encode point sets through a PH Encoder, apply self-attention within each homology dimension, model inter-dimensional relationships via bidirectional cross-attention, aggregate features with max and mean pooling, and project to the final representation $\mathbf{t} \in \mathbb{R}^{256}$.

To process PDs, we adopt a PointNet-like network~\cite{qi2017pointnet}, called the PH Encoder. Since PDs have variable sizes, we select the top-$k$ most persistent features ($k_{H_0} = 48$, $k_{H_1} = 96$), which capture nearly all significant topological structures across our benchmarks. We represent each selected birth-death pair as $\mathbf{p}_i = (b_i, d_i)$, where $i \in \{1,\ldots,144\}$ indexes the sequence of $H_0$ and $H_1$ features. To enable the encoder to distinguish between homology dimensions, we augment each pair with one-hot encoding: $\mathbf{en}_i = [1, 0]$ for the first 48 points ($H_0$) and $\mathbf{en}_i = [0, 1]$ for the remaining 96 points ($H_1$), giving $\tilde{\mathbf{p}}_i = [\mathbf{p}_i; \mathbf{en}_i] \in \mathbb{R}^4$. The PH Encoder processes all 144 $\tilde{\mathbf{p}}_i$ independently through four fully-connected layers (4→64→128→256→384). We then separate the encoded features into $\mathbf{h}_0 \in \mathbb{R}^{48 \times 384}$ and $\mathbf{h}_1 \in \mathbb{R}^{96 \times 384}$ for the two homology dimensions.

Given the embeddings $\mathbf{h}_0$ and $\mathbf{h}_1$ from the PH Encoder, we effectively learn topological representations that capture structural patterns. Medical image analysis commonly relies on two key characteristics. (1) Certain anatomical regions are more medically significant than others. For example, in tissue analysis, tumor regions are more diagnostically relevant than background tissues~\cite{litjens2017survey}. This requires our model to distinguish the importance of topological features within each homology dimension (e.g., $H_0$ features representing tumor regions vs background tissues). (2) Structural relationships among different types of anatomical features provide critical diagnostic information. For example, glandular lumens occurring within tumor regions indicate malignancy, while similar lumens within healthy tissues are benign~\cite{qaiser2019fast}. Thus, our model should capture geometric dependencies between $H_0$ features (e.g., representing tumor regions or healthy tissues) and $H_1$ features (e.g., representing glandular lumens). We address both requirements through hierarchical attention mechanisms: self-attention within each dimension distinguishes feature importance, and cross-attention between dimensions captures geometric dependencies.

Specifically, as shown in Fig.~\ref{fig:overview}(b), we first apply self-attention within each homology dimension with weighted residual connections:
$
\mathbf{h}'_0 = \mathbf{h}_0 + \lambda_0 \cdot \text{SelfAttn}(\mathbf{h}_0), \, \mathbf{h}'_1 = \mathbf{h}_1 + \lambda_1 \cdot \text{SelfAttn}(\mathbf{h}_1),
$
where $\lambda_0 = \lambda_1 = 0.5$ balance learned attention with the original features. Bidirectional cross-attention then models geometric relationships:
$
\mathbf{h}^{\leftrightarrow}_0 = \text{CrossAttn}(\mathbf{h}'_0, \mathbf{h}'_1), \, \mathbf{h}^{\leftrightarrow}_1 = \text{CrossAttn}(\mathbf{h}'_1, \mathbf{h}'_0),
$
where $H_0$ features serve as query and $H_1$ as key and value in $\mathbf{h}^{\leftrightarrow}_0$, and vice versa in $\mathbf{h}^{\leftrightarrow}_1$, modeling topological dependencies such as hole containment within components. We concatenate the cross-attention outputs as $\mathbf{h}_{\text{cross}} = [\mathbf{h}^{\leftrightarrow}_0; \mathbf{h}^{\leftrightarrow}_1] \in \mathbb{R}^{144 \times 384}$, {where [$\cdot$;$\cdot$] denotes concatenation} (Cross-Attn in Fig.~\ref{fig:overview}(b)). Both self-attention and cross-attention use 4-head multi-head attention.

We aggregate features from self-attention ($\mathbf{h}'_0$, $\mathbf{h}'_1$) and cross-attention ($\mathbf{h}_{\text{cross}}$) using max and mean pooling. Max pooling captures the most salient topological structures, while mean pooling preserves global distributional information. Applying both pooling operations to each of these three feature sets yields six pooled vectors of 384 dimensions each, denoted as $\mathbf{h}_{\text{pool}} \in \mathbb{R}^{6 \times 384}$ (Max \& Mean Pooling in Fig.~\ref{fig:overview}(b)). A projection head (a three-layer MLP: 2304→768→512→256) then maps the flattened $\mathbf{h}_{\text{pool}}$ to the final representation $\mathbf{t} \in \mathbb{R}^{256}$ (Proj Head in Fig.~\ref{fig:overview}(b)). {During pretraining, the H-Topo.~Encoder is trained with the contrastive loss $\mathcal{L}_{\mathcal{M}}$.}
% During pretraining (Section~\ref{sec:overview}), the H-Topo.~Encoder is trained independently using the contrastive objective defined in Eq.~\eqref{eq:topo_pretrain}.

\begin{table}[t]
\centering
\caption{Dataset statistics for the five medical image datasets.}
\label{tab:datasets}
\scriptsize
\begin{tabular}{llcc}
\toprule
\textbf{Dataset} & \textbf{Modality} & \textbf{Classes} & \textbf{Train / Test} \\
\midrule
PathMNIST & Colon Pathology & 9 & 89,996 / 7,180 \\
OCTMNIST & Retinal OCT & 4 & 97,477 / 1,000 \\
OrganSMNIST & Abdominal CT (Sagittal) & 11 & 13,940 / 8,829 \\
ISIC2019 & Skin Lesion Dermoscopy & 8 & 25,331 / 8,238 \\
Kvasir & GI Endoscopy & 8 & 3,200 / 800 \\
\bottomrule
\end{tabular}
\vspace{-0.4cm}
\end{table}

\begin{table*}[t]
\centering
\scriptsize
\caption{Performance comparison across five medical image datasets. The results show mean $\pm$ standard deviation over five independent runs. Statistical significance is assessed via paired t-tests: $^{***}$: $p<0.001$, $^{**}$: $p<0.01$, $^{*}$: $p<0.05$. $^{\dagger}$ indicates 
%significant 
performance decrease. The Avg.~columns report dataset-averaged metric values.}
\label{tab:main_results}
\setlength{\tabcolsep}{0.8pt}
\begin{tabular}{l*{12}{c}}
\toprule
& \multicolumn{2}{c}{\textbf{Path}} & \multicolumn{2}{c}{\textbf{OrganS}} & \multicolumn{2}{c}{\textbf{OCT}} & \multicolumn{2}{c}{\textbf{ISIC}} & \multicolumn{2}{c}{\textbf{Kvasir}} & \multicolumn{2}{c}{\textbf{Avg.}} \\
\cmidrule(lr){2-3}\cmidrule(lr){4-5}\cmidrule(lr){6-7}\cmidrule(lr){8-9}\cmidrule(lr){10-11}\cmidrule(lr){12-13}
\textbf{Method} & ACC & AUC & ACC & AUC & ACC & AUC & ACC & AUC & ACC & AUC & ACC & AUC \\
\midrule
SimCLR~\cite{chen2020simple} & $92.57_{\pm 0.31}$ & $99.12_{\pm 0.11}$ & $77.12_{\pm 0.26}$ & $96.41_{\pm 0.26}$ & $69.41_{\pm 0.56}$ & $91.83_{\pm 0.43}$ & $66.04_{\pm 2.29}$ & $88.01_{\pm 1.11}$ & $74.33_{\pm 1.28}$ & $95.89_{\pm 0.76}$ & $75.89_{\pm 0.51}$ & $94.25_{\pm 0.11}$ \\
\quad +TopoCL & $93.21_{\pm 0.14}^*$ & $99.28_{\pm 0.05}$ & $78.62_{\pm 0.06}^{***}$ & $96.75_{\pm 0.33}^*$ & $73.10_{\pm 0.21}^{***}$ & $93.41_{\pm 0.26}^{**}$ & $71.62_{\pm 0.21}^{**}$ & $89.82_{\pm 0.77}^{**}$ & $78.61_{\pm 0.41}^{**}$ & $97.11_{\pm 0.32}^*$ & $79.03_{\pm 0.10}^{***}$ & $95.27_{\pm 0.15}^{***}$ \\
\midrule
MoCo-v3~\cite{chen2021mocov3} & $93.13_{\pm 0.42}$ & $99.07_{\pm 0.09}$ & $78.02_{\pm 0.34}$ & $99.57_{\pm 0.31}$ & $80.02_{\pm 0.34}$ & $96.01_{\pm 0.04}$ & $74.98_{\pm 2.01}$ & $92.86_{\pm 1.12}$ & $88.42_{\pm 2.01}$ & $97.37_{\pm 2.01}$ & $82.91_{\pm 0.66}$ & $96.98_{\pm 0.60}$ \\
\quad +TopoCL & $94.55_{\pm 0.20}^{**}$ & $99.24_{\pm 0.09}^{**}$ & $80.58_{\pm 0.09}^{***}$ & $98.75_{\pm 0.21}^{\dagger}$ & $82.09_{\pm 0.26}^{***}$ & $97.09_{\pm 0.13}^{***}$ & $78.44_{\pm 0.83}^*$ & $93.26_{\pm 0.97}$ & $91.17_{\pm 0.83}^*$ & $98.81_{\pm 0.46}$ & $85.37_{\pm 0.11}^{***}$ & $97.43_{\pm 0.27}$ \\
\midrule
BYOL~\cite{grill2020bootstrap} & $93.42_{\pm 0.09}$ & $99.31_{\pm 0.14}$ & $76.41_{\pm 0.19}$ & $97.33_{\pm 0.08}$ & $76.02_{\pm 0.26}$ & $95.89_{\pm 0.25}$ & $73.82_{\pm 3.19}$ & $90.75_{\pm 0.79}$ & $85.67_{\pm 0.98}$ & $96.91_{\pm 0.13}$ & $81.07_{\pm 0.49}$ & $96.04_{\pm 0.20}$ \\
\quad +TopoCL & $94.89_{\pm 0.19}^{***}$ & $99.41_{\pm 0.06}$ & $79.53_{\pm 0.12}^{***}$ & $98.03_{\pm 0.21}^{**}$ & $79.13_{\pm 0.24}^{***}$ & $96.18_{\pm 0.16}^*$ & $77.18_{\pm 0.62}$ & $91.52_{\pm 0.46}$ & $88.83_{\pm 0.88}^{**}$ & $97.55_{\pm 0.48}^*$ & $83.91_{\pm 0.14}^{***}$ & $96.54_{\pm 0.20}^{*}$ \\
\midrule
DINO~\cite{caron2021emerging} & $91.92_{\pm 0.15}$ & $98.99_{\pm 0.18}$ & $71.11_{\pm 0.31}$ & $95.39_{\pm 0.08}$ & $61.21_{\pm 0.33}$ & $91.27_{\pm 0.16}$ & $63.24_{\pm 2.45}$ & $85.28_{\pm 0.89}$ & $71.42_{\pm 1.86}$ & $96.99_{\pm 0.27}$ & $71.78_{\pm 0.50}$ & $93.58_{\pm 0.19}$ \\
\quad +TopoCL & $94.57_{\pm 0.08}^{***}$ & $99.38_{\pm 0.05}^{**}$ & $77.21_{\pm 0.17}^{***}$ & $96.96_{\pm 0.13}^{***}$ & $65.03_{\pm 0.19}^{***}$ & $93.87_{\pm 0.15}^{***}$ & $67.08_{\pm 0.17}^*$ & $88.38_{\pm 0.59}^{**}$ & $77.99_{\pm 1.27}^{**}$ & $98.31_{\pm 0.25}^{**}$ & $76.38_{\pm 0.30}^{***}$ & $95.38_{\pm 0.17}^{***}$ \\
\midrule
Barlow~\cite{zbontar2021barlow} & $93.09_{\pm 0.11}$ & $99.41_{\pm 0.06}$ & $76.76_{\pm 0.26}$ & $98.13_{\pm 0.09}$ & $78.08_{\pm 0.49}$ & $96.91_{\pm 0.15}$ & $62.70_{\pm 1.41}$ & $87.11_{\pm 0.26}$ & $73.41_{\pm 1.29}$ & $97.57_{\pm 0.22}$ & $76.81_{\pm 0.37}$ & $95.83_{\pm 0.08}$ \\
\quad +TopoCL & $94.91_{\pm 0.05}^{***}$ & $99.52_{\pm 0.04}$ & $79.69_{\pm 0.09}^{***}$ & $98.75_{\pm 0.11}^{***}$ & $80.51_{\pm 0.28}^{***}$ & $97.28_{\pm 0.18}^*$ & $67.22_{\pm 0.23}^{**}$ & $89.02_{\pm 0.45}^{**}$ & $78.08_{\pm 1.08}^{**}$ & $98.13_{\pm 0.41}^*$ & $80.08_{\pm 0.20}^{***}$ & $96.54_{\pm 0.08}^{***}$ \\
\bottomrule
\end{tabular}
\vspace{-0.4cm}
\end{table*}

\subsection{Mixture-of-Experts Fusion}
\label{sec:moe}

After independently pretraining the visual and topology encoders (Section~\ref{sec:overview}), we jointly fine-tune them with a fusion module using contrastive learning. Existing fusion strategies include concatenation~\cite{ngiam2011multimodal}, gated fusion~\cite{arevalo2017gated}, cross-attention~\cite{vaswani2017attention}, etc., each assuming that a single strategy is optimal for all samples. However, medical images often exhibit substantial heterogeneity: texture-dominated samples (e.g., in dermoscopy images, pigmentation patterns are diagnostic) may benefit more from visual features, while structure-dominated samples (e.g., in histopathology, glandular connectivity is critical) require stronger topological signals~\cite{litjens2017survey}. Such diversities motivate the use of a Mixture-of-Experts (MoE) approach~\cite{shazeer2017outrageously}, which combines multiple fusion strategies rather than committing to a single strategy. We design five fusion experts covering diverse paradigms (vision-only, topology-only, concatenation, gated blending, and cross-modal interaction), enabling the model to learn sample-specific expert weights.

Fig.~\ref{fig:overview}(a) illustrates the joint fine-tuning workflow with MoE fusion. Given an input image $x$, we apply topology-aware augmentations (Section~\ref{sec:topo_aug}) to generate two views $x_{\text{topo}}^w$ and $x_{\text{topo}}^s$, and compute their PDs $\text{topo}^w = \text{PD}(x_{\text{topo}}^w)$ and $\text{topo}^s = \text{PD}(x_{\text{topo}}^s)$. The two pretrained encoders $f_v$ and $f_t$ process the augmented images and PDs, respectively. We project both encoder outputs to a shared 256-dimensional embedding space through separate 3-layer MLP projection heads, yielding visual features $\mathbf{f}^w, \mathbf{f}^s$ and topological features $\mathbf{t}^w, \mathbf{t}^s$ (Vis.-Topo.~Embed in Fig.~\ref{fig:overview}(a)). A multi-gating network processes the concatenated features $[\mathbf{f}; \mathbf{t}]$ to compute sample-specific weights, $\text{gate}^w, \text{gate}^s \in \mathbb{R}^5$. In parallel, the five fusion experts process the features based on their respective strategies (detailed below). The expert outputs are combined through weighted summation using the gate weights (Weighted Sum in Fig.~\ref{fig:overview}(a)), and a projection head produces the final embeddings $\mathbf{z}^w, \mathbf{z}^s \in \mathbb{R}^{256}$.

To enable this sample-adaptive fusion, we design five expert networks with distinct fusion strategies. For clarity, we describe the operations for a single view; the same approach applies to both views. All experts have access to both projected visual features $\mathbf{f}$ and topological features $\mathbf{t}$ (corresponding to $\mathbf{f}^w, \mathbf{t}^w$ or $\mathbf{f}^s, \mathbf{t}^s$), but process them with their own strategies. {\bf Expert 1}, Vis.-Only ($\mathbf{e}_1$), and {\bf Expert 2}, Topo.-Only ($\mathbf{e}_2$), process single feature types:
$
\mathbf{e}_1 = \text{MLP}(\mathbf{f}), \, \mathbf{e}_2 = \text{MLP}(\mathbf{t}),
$
allowing the model to rely solely on visual or topological information. {\bf Expert 3}, Concat ($\mathbf{e}_3$), processes concatenated features:
$
\mathbf{e}_3 = \text{MLP}([\mathbf{f}; \mathbf{t}]),
$
capturing additive interactions. {\bf Expert 4}, Gated Blending ($\mathbf{e}_4$), learns sample-specific blending:
$
\mathbf{g} = \sigma(\text{MLP}([\mathbf{f}; \mathbf{t}])), \, \mathbf{e}_4 = \text{MLP}(\mathbf{g} \odot \mathbf{f} + (1-\mathbf{g}) \odot \mathbf{t}),
$
where $\sigma$ is sigmoid and $\odot$ denotes element-wise multiplication. {\bf Expert 5}, Cross-Attn ($\mathbf{e}_5$), applies bidirectional cross-attention with residual connections:
$
\mathbf{f}' = \mathbf{f} + \text{CrossAttn}(Q=\mathbf{f}, KV=\mathbf{t}), \,
\mathbf{t}' = \mathbf{t} + \text{CrossAttn}(Q=\mathbf{t}, KV=\mathbf{f}), \,
\mathbf{e}_5 = \text{MLP}([\mathbf{f}'; \mathbf{t}']),
$
where $Q$ denotes query and $KV$ denotes key-value pairs in the attention mechanism. All experts use 3-layer MLPs and output 512-dimensional representations $\mathbf{e}_i \in \mathbb{R}^{512}$.

These expert outputs are combined with a gating mechanism. A multi-gating network (a 3-layer MLP: 512→256→128→5) takes concatenated features $[\mathbf{f}; \mathbf{t}]$ as input and produces normalized weights:
$
\text{gate} = \text{softmax}(\text{Multi-Gating}([\mathbf{f}; \mathbf{t}])) \in \mathbb{R}^5.
$
The fused representation is obtained via weighted sum:
$
\mathbf{h} = \sum_{i=1}^{5} \text{gate}_i \cdot \mathbf{e}_i \in \mathbb{R}^{512},
$
which a projection head then maps to $\mathbf{z} \in \mathbb{R}^{256}$ (Proj Head in Fig.~\ref{fig:overview}(a)). During joint fine-tuning, both views produce embeddings $\mathbf{z}^w, \mathbf{z}^s$, optimized using the contrastive objective in Section~\ref{sec:overview}. During inference, given a single image, we extract its ROIs, compute the PD without augmentations, process the image through $f_v$ and the PD through $f_t$ respectively, and fuse the projected features through the gating mechanism and experts to produce the final representation for downstream tasks.

\section{Experiments}
\label{sec:experiments}

\noindent\textbf{Datasets.} We evaluate on five medical image classification datasets covering diverse modalities and anatomical regions (Table~\ref{tab:datasets}): PathMNIST, OCTMNIST, OrganSMNIST, ISIC2019, and Kvasir. All images are preprocessed to $224 \times 224$ following MedMNIST v2 protocols~\cite{medmnistv2}.

\noindent\textbf{Baseline Methods.} We integrate TopoCL with five representative CL methods: SimCLR~\cite{chen2020simple}, MoCo-v3~\cite{chen2021mocov3}, BYOL~\cite{grill2020bootstrap}, DINO~\cite{caron2021emerging}, and Barlow Twins~\cite{zbontar2021barlow}. All methods use ResNet-50 as the visual encoder backbone and are trained for 150 epochs.

\noindent\textbf{Evaluation Protocol.} Following standard self-supervised evaluation~\cite{chen2020simple,caron2021emerging}, we assess representation quality via linear probing: a linear classifier is trained on frozen encoder features for 100 epochs and evaluated on the test set. We report accuracy (ACC) and macro-averaged AUC, averaged over five runs with statistical significance via paired t-tests.

\noindent\textbf{Implementation Details.} We apply SAM-ViT-H~\cite{kirillov2023segment} to extract ROIs and compute PDs with GUDHI~\cite{gudhi:CubicalComplex}, retaining top-48 $H_0$ and top-96 $H_1$ features. The Hierarchical Topology Encoder uses 4-head attention with dimension 384. Models are trained on H100 GPUs with batch size 256, AdamW optimizer (learning rate $3 \times 10^{-4}$), and cosine annealing. Full details are in Supplementary Material.

\subsection{Main Results}
Table~\ref{tab:main_results} presents comprehensive results on integrating our TopoCL with five CL methods and five medical image datasets. Averaged across all datasets, TopoCL consistently improves the baseline performance, with ACC gains ranging from +2.46\% to +4.60\% and AUC gains from +0.45\% to +1.80\%, giving overall average improvements of +3.26\% ACC and +0.90\% AUC. Statistical analysis via paired t-tests shows strong statistical significance: 86\% (43/50) of individual dataset-metric comparisons and 90\% (9/10) of dataset-averaged metrics achieve statistical significance (p$<$0.05), with 80\% of the averaged metrics reaching p$<$0.001. Importantly, observe that TopoCL benefits all five CL methods, with 
every baseline showing positive average improvements in both ACC and AUC.

TopoCL yields particularly strong results with DINO, achieving the largest average gains (+4.60\% ACC, +1.80\% AUC) with statistical significance across all ten dataset-metric combinations. Dataset-wise, OCT and Kvasir show substantial improvements (average gains of +3.02\% and +4.29\% ACC, respectively), with beneficial structural patterns from topological features. ISIC exhibits higher variance due to skin lesion heterogeneity, yielding some moderate improvements despite consistent positive trends.

One notable exception occurs with MoCo-v3 on OrganSMNIST, where AUC decreases from 99.57\% to 98.75\% (p$=$0.023) despite a considerable ACC improvement (+2.56\%, p$<$0.001). This suggests a trade-off in the near-saturated situations (baseline AUC$>$99.5\%), where topological features help optimize classification 
%boundaries 
performance at the expense of ranking confidence. Nevertheless, these results validate that TopoCL provides robust and statistically significant improvements across diverse CL methods and medical imaging modalities.

\subsection{Topology-Aware Augmentation Design Choices}
\begin{table}[bt]
\centering
\scriptsize
\caption{Ablation of topology-aware augmentation configurations using MoCo-v3+TopoCL. 
%The rows compare standard visual augmentations, topology-aware augmentations on full images vs. ROIs, and symmetric vs. mixed topology-aware augmentation pairs. 
ACC results (\%) are reported.} %Baseline aug: MoCo-v3 default augmentations. Topo: topology-aware augmentation with SAM-extracted ROI and topology-weak+topology-strong pairs by default. Full image: exception where persistent diagrams computed on entire image instead of ROI. Weak+weak / strong+strong: symmetric augmentation pairs. ACC (\%) reported.}
\label{tab:topo_aug_ablation}
\setlength{\tabcolsep}{4pt}
\begin{tabular}{lccc}
\toprule
\textbf{Configuration} & \textbf{OrganS} & \textbf{ISIC} & \textbf{Kvasir} \\
\midrule
TopoCL (weak+strong) & 75.51$_{\pm 1.56}$ & 72.92$_{\pm 2.32}$ & 84.63$_{\pm1.85}$\\
TopoCL (full image) & 76.34$_{\pm 1.46}$ & 74.35$_{\pm 1.58}$ & 87.41$_{\pm 1.88}$\\
TopoCL (topo-weak+topo-weak) & 78.98$_{\pm 0.37}$& 76.39$_{\pm 0.55}$ & 90.01$_{\pm 0.28}$ \\
TopoCL (topo-strong+topo-strong) & 78.02$_{\pm 0.68}$ & 75.99$_{\pm 1.02}$ & 89.64$_{\pm 0.91}$\\
\midrule
\textbf{TopoCL (topo-weak+topo-strong)} & \textbf{80.58$_{\pm0.99}$} & \textbf{78.44$_{\pm0.83}$} & \textbf{91.17$_{\pm0.43}$} \\
\bottomrule
\end{tabular}
\vspace{-0.5cm}
\end{table}
%\subsection{Topology-Aware Augmentation Analysis}

We evaluate topology-aware augmentation design choices using MoCo-v3+TopoCL on three datasets. We compare five configurations:
(a) \emph{weak+strong}: standard visual weak/strong augmentations (as in the 
baseline MoCo-v3);
(b) \emph{full image}: topology-aware weak/strong augmentations with PDs 
computed on the entire images instead of SAM-extracted ROIs;
(c) \emph{topo-weak+topo-weak} and (d) \emph{topo-strong+topo-strong}: 
symmetric topology-aware augmentation pairs (both views use the same 
augmentation strength) on SAM-extracted ROIs;
(e) \emph{topo-weak+topo-strong}: mixed topology-aware augmentation pairs 
(different augmentation strengths) on SAM-extracted ROIs.

In Table~\ref{tab:topo_aug_ablation}, we observe:
(1) Replacing standard visual augmentations with topology-aware augmentations yields substantial gains; the mixed \emph{topo-weak+topo-strong} pairing achieves the best results across the datasets. This supports our premise that explicitly controlling topological perturbations preserves semantically meaningful structures.
(2) Computing PDs on SAM-extracted ROIs outperforms computing them on full images, indicating that focusing topology measurements on relevant anatomy (rather than background) enables more precise topological control.
(3) Symmetric topology-aware pairs (\emph{topo-weak+topo-weak} or 
\emph{topo-strong+topo-strong}) improve over standard visual augmentations, but the mixed pairing provides the most informative contrast, likely because it exposes the encoder to controlled yet complementary topology variations.

% Table~\ref{tab:topo_aug_ablation} ablates topology-aware augmentation design choices on MoCo-v3+TopoCL across three datasets. Compared to standard augmentations, topology-aware augmentation with full images provides consistent but modest improvements (+0.8-2.8\% ACC), validating that controlling topological perturbations via bottleneck distance preserves semantically meaningful topology. ROI extraction substantially boosts performance, with Topo(weak+weak) outperforming Topo(full image) by +2.0-2.6\% across datasets. This confirms that SAM-based region extraction focuses topology computation on relevant anatomical structures rather than background, enabling more precise topological control.

% Among augmentation strength combinations, weak+strong pairs achieve the best performance (80.58\%, 78.44\%, 91.17\%), outperforming both symmetric configurations. The weak+weak setting underperforms due to insufficient diversity between contrastive views, while strong+strong pairs risk over-distorting topology and destroying semantic content. The asymmetric weak+strong strategy balances these concerns, providing diverse views while maintaining structural integrity. These results validate our topology-aware augmentation framework: SAM ROI extraction combined with asymmetric weak+strong calibration provides effective topology-preserving data augmentation for medical image contrastive learning.

\subsection{Ablation Studies}
\begin{table}[bt]
\centering
\scriptsize
\caption{Ablation 
%study 
with SimCLR+TopoCL on three 
%representative 
datasets. We 
%systematically 
evaluate: (1) pretraining strategies (rows 1-3); (2) hierarchical topology encoder components (rows 4-6); (3) MoE expert choices (rows 7-11). ACCs (\%) averaged over five runs are reported. 
%w/o = without. Abbreviations: vis.=visual, topo.=topological, attn=attention, concat=concatenation, hier.=hierarchical.
}
\label{tab:ablation}
\setlength{\tabcolsep}{5pt}
\begin{tabular}{lccc}
\toprule
\textbf{Configuration} & \textbf{Path} & \textbf{OrganS} &  \textbf{OCT}\\
\midrule
w/o pretraining & $68.98_{\pm 1.43}$ & $50.61_{\pm 0.87}$ & $48.72_{\pm 1.77}$ \\
w/o vis. pretrain & $81.53_{\pm 0.98}$ & $64.72_{\pm 1.21}$ & $66.98_{\pm 0.83}$ \\
w/o topo. pretrain & $87.64_{\pm 0.92}$ & $72.56_{\pm 1.11}$ & $69.37_{\pm 0.89}$\\
\midrule
w/o hier. attn & $86.51_{\pm 0.93}$ & $72.85_{\pm 0.76}$ & $67.62_{\pm 1.03}$\\
w/o self-attn & $91.24_{\pm 0.85}$ & $76.03_{\pm 0.56}$ & $70.28_{\pm 0.66}$ \\
w/o cross-attn & $87.59_{\pm 0.85}$ & $74.66_{\pm 0.71}$ & $69.38_{\pm 0.57}$\\
\midrule
w/o vis-only expert & $92.59_{\pm 0.43}$  & $77.47_{\pm 0.51}$ &  $71.85_{\pm 0.34}$\\
w/o topo-only expert & $93.01_{\pm 0.18}$  & $78.15_{\pm 0.11}$ & $72.53_{\pm 0.25}$ \\
w/o concat expert & $92.88_{\pm 0.58}$ & $78.05_{\pm 0.43}$ & $72.68_{\pm 0.41}$ \\
w/o gated expert & $92.05_{\pm 0.56}$ & $76.89_{\pm 0.47}$ & $71.63_{\pm 0.66}$\\
w/o cross-attn expert & $91.75_{\pm 0.48}$ & $76.03_{\pm 0.56}$ & $70.95_{\pm 0.48}$ \\
\midrule
Full model & $93.21_{\pm 0.14}$ & $78.62_{\pm 0.06}$ &  $73.10_{\pm 0.21}$\\
\bottomrule
\end{tabular}
\vspace{-0.4cm}
\end{table}

\begin{figure*}[bt]
    \centering
\includegraphics[width=0.65\linewidth,trim={0cm 0.1cm 0.0cm 0.0cm},clip]{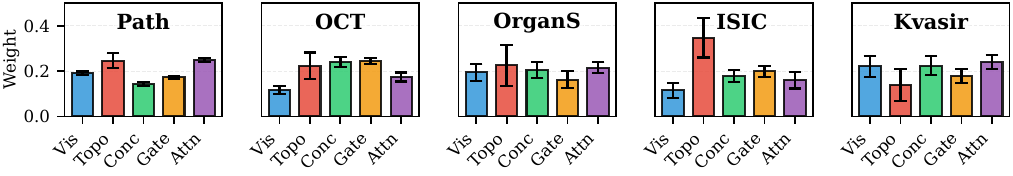}

\caption{Expert gating analysis for TopoCL integrated with BYOL on five datasets. Error bars show standard deviation across test samples. Five experts: visual-only (Vis), topology-only (Topo), concatenation (Conc), gated blending (Gate), and cross-attention (Attn).}
\label{fig:expert_gating}
\vspace{-0.4cm}
\end{figure*}

To validate the effect of each design component, we conduct systematic ablations with SimCLR+TopoCL on three datasets (Table~\ref{tab:ablation}). We evaluate the pretraining strategies, topology encoder architecture, and MoE expert choices.

\noindent\textbf{Pretraining Strategies.} Visual and topological pretraining are both essential; their 
%complete 
removal causes severe degradation (68.98\% vs 93.21\% on PathMNIST). Visual pretraining has a much larger impact (performance drops to 81.53\%) than topological pretraining (87.64\%), confirming that visual features provide primary representations and topological features offer complementary structural information.

\noindent\textbf{Topology Encoder Architecture.} The hierarchical design is critical; its complete removal drops the performance by 6.70\% on PathMNIST. Cross-attention (87.59\%) is more important than self-attention (91.24\%), as it captures relationships between connected components ($H_0$) and 
%topological 
holes ($H_1$), validating our hierarchical integration strategy.

\noindent\textbf{MoE Expert Effects.} All five experts contribute to the performance, with fusion-based experts being most effective: Cross-attention expert (1.73\% average drop when removed), gated expert (1.45\%), visual-only expert (1.01\%), concat expert (0.44\%), and topology-only expert (0.41\%). The finding of fusion experts outweighing feature-specific experts validates our MoE design, in which topological information is effectively leveraged when adaptively integrated with visual features rather than used in isolation.

\subsection{Computational Cost Analysis}

\begin{table}[bt]
\centering
\scriptsize
\caption{Computational cost comparison on ISIC2019. 
%Tput. = throughput (samples/sec).
}
%Our topology-aware framework improves all contrastive learning methods with modest overhead.}
\label{tab:method_comparison}
\setlength{\tabcolsep}{1pt}
\begin{tabular}{lccccc}
\toprule
\textbf{Method} & \textbf{Train (h)} & \textbf{Params (M)} & \textbf{FLOPs (G)} & \textbf{Tput.} & \textbf{Overhead} \\
\midrule
SimCLR & 1.58 & 24.89 & 4.11 & 668 & - \\
\quad +TopoCL & 1.84 & 31.08 & 6.11 & 574 & +16\% \\
BYOL & 1.72 & 53.28 & 2.88 & 614 & - \\
\quad +TopoCL & 1.96 & 60.48 & 4.58 & 539 & +14\% \\
DINO & 1.79 & 45.16 & 5.23 & 590 & - \\
\quad +TopoCL & 1.99 & 51.00 & 7.03 & 531 & +11\% \\
Barlow & 1.28 & 24.89 & 4.57 & 825 & - \\
\quad +TopoCL & 1.51 & 31.08 & 6.54 & 700 & +18\% \\
MoCo-v3 & 1.72 & 52.89 & 3.09 & 614 & - \\
\quad +TopoCL & 1.83 & 59.73 & 5.33 & 577 & +6\% \\
\bottomrule
\end{tabular}
\vspace{-0.4cm}
\end{table}

\begin{table}[bt]
\centering
\scriptsize
\caption{Training and inference cost breakdown for SimCLR+TopoCL on ISIC2019.}
\label{tab:simclr_breakdown}
\setlength{\tabcolsep}{4pt}
\begin{tabular}{lccccc}
\toprule
\multirow{2}{*}{\textbf{Component}} & \multicolumn{2}{c}{\textbf{Training}} & \multicolumn{2}{c}{\textbf{Inference}} \\
\cmidrule(lr){2-3} \cmidrule(lr){4-5}
& \textbf{Time (h)} & \textbf{\%} & \textbf{Params (M)} & \textbf{FLOPs (G)} \\
\midrule
Visual Encoder & 1.06 & 57.61 & 24.03 & 4.11 \\
Topo Encoder & 0.35 & 19.02 & 3.83 & 1.99 \\
MoE+Proj & 0.43 & 23.37 & 3.23 & 0.01 \\
\midrule
\textbf{Total} & \textbf{1.84} & \textbf{100.0} & \textbf{31.08} & \textbf{6.11} \\
\bottomrule
\end{tabular}
\vspace{-0.4cm}
\end{table}

Table~\ref{tab:method_comparison} analyzes the computational overhead of TopoCL with the five CL methods. Integrating topology-aware augmentation and MoE fusion adds modest overhead: Training time increases by 6-18\%, depending on the method (13\% average), with SimCLR+TopoCL completing 150 epochs in 1.84 hours on H100. The parameter count increases by 6.5M on average (17\%), while FLOPs increase by $\sim$2 GFLOPs (51\% average increase). The throughput reduction averages 12\%, which is acceptable for medical imaging applications where training is typically conducted once.

Table~\ref{tab:simclr_breakdown} provides cost breakdown for SimCLR+TopoCL. The Visual Encoder (24.03M parameters) refers to the ResNet-50 backbone shared with the baseline SimCLR. The baseline's parameter count of 24.89M in Table~\ref{tab:method_comparison} includes an additional 0.86M projection head, which TopoCL replaces with the Topo Encoder (3.83M) and MoE+Proj (3.23M) modules. The visual encoder dominates both training time (57.61\%) and inference cost (67\% of FLOPs), while the topology encoder contributes 19\% of training time and 33\% of inference FLOPs. For deployment, topology features can be precomputed offline (0.64 h for ISIC2019), reducing inference overhead to primarily the MoE+Proj component and making TopoCL practical for clinical applications.

\subsection{Expert Gating Analysis}

Fig.~\ref{fig:expert_gating} reveals dataset-specific gating patterns for BYOL. PathMNIST strongly prefers topology-only and cross-attention experts, indicating histopathology benefits from both standalone topological features and their integration with visual features. OCTMNIST and OrganSMNIST exhibit balanced gating, indicating retinal OCT and abdominal CT structures require diverse fusion strategies. ISIC2019 demonstrates the strongest topology-only preference, likely because skin lesion boundaries encoded by PH are particularly discriminative. Kvasir shows balanced weights with a slight emphasis on concatenation and visual-only experts.

The topology-only expert receives varying importance across the datasets (high on ISIC2019, lower on the others), demonstrating adaptive determination of when standalone topological features are most valuable vs when integration through fusion experts is preferred. While Table~\ref{tab:main_results} shows that topological features improve all methods, this analysis reveals that an optimized fusion strategy is dataset-dependent. The multi-gating network learns these preferences without supervision, validating TopoCL's adaptive design. Similar patterns for other CL methods (see Supplementary Material) demonstrate broad generalization.
\section{Conclusions}
\label{sec:conclusion}
In this paper, we introduced TopoCL, a new general topological contrastive learning framework for medical imaging that augments standard contrastive learning with explicit topological preservation. Our method combines topology-aware augmentations calibrated via relative bottleneck distance of persistence diagrams, a Hierarchical Topology Encoder that captures both $H_0$ and $H_1$ topological features, and an adaptive Mixture-of-Experts module that fuses visual and topological representations. Extensive experiments on five medical image benchmarks using five representative CL baselines (SimCLR, MoCo-v3, BYOL, DINO, and Barlow Twins) showed that TopoCL improves the average accuracy by +3.26\% with strong statistical significance, confirming that topological cues provide complementary and robust supervisory signals.
{
    \small
    \bibliographystyle{ieeenat_fullname}
    \bibliography{main}

@String(AAAI = {AAAI})

@article{litjens2017survey,
  title={A survey on deep learning in medical image analysis},
  author={Litjens, Geert and Kooi, Thijs and Bejnordi, Babak Ehteshami and Setio, Arnaud Arindra Adiyoso and Ciompi, Francesco and Ghafoorian, Mohsen and Van Der Laak, Jeroen Awm and Van Ginneken, Bram and S{\'a}nchez, Clara I},
  journal={Medical Image Analysis},
  volume={42},
  pages={60--88},
  year={2017},
  publisher={Elsevier}
}

@inproceedings{irvin2019chexpert,
  title={{CheXpert}: A large chest radiograph dataset with uncertainty labels and expert comparison},
  author={Irvin, Jeremy and Rajpurkar, Pranav and Ko, Michael and Yu, Yifan and Ciurea-Ilcus, Silviana and Chute, Chris and Marklund, Henrik and Haghgoo, Behzad and Ball, Robyn and Shpanskaya, Katie and others},
  booktitle={Proceedings of the AAAI Conference on Artificial Intelligence},
  volume={33},
  number={01},
  pages={590--597},
  year={2019}
}

@article{zhou2024comprehensive,
  title={A comprehensive survey on deep clustering: Taxonomy, challenges, and future directions},
  author={Zhou, Sheng and Xu, Hongjia and Zheng, Zhuonan and Chen, Jiawei and Li, Zhao and Bu, Jiajun and Wu, Jia and Wang, Xin and Zhu, Wenwu and Ester, Martin},
  journal={ACM Computing Surveys},
  volume={57},
  number={3},
  pages={1--38},
  year={2024},
  publisher={ACM New York, NY}
}

@inproceedings{chen2020simple,
  title={A simple framework for contrastive learning of visual representations},
  author={Chen, Ting and Kornblith, Simon and Norouzi, Mohammad and Hinton, Geoffrey},
  booktitle={Proceedings of the International Conference on Machine Learning},
  pages={1597--1607},
  year={2020}
}

@article{rajpurkar2017chexnet,
  title={{CheXNet}: Radiologist-level pneumonia detection on chest {X-rays} with deep learning},
  author={Rajpurkar, Pranav and Irvin, Jeremy and Zhu, Kaylie and Yang, Brandon and Mehta, Hershel and Duan, Tony and Ding, Daisy and Bagul, Aarti and Langlotz, Curtis and Shpanskaya, Katie and others},
  journal={arXiv preprint arXiv:1711.05225},
  year={2017}
}

@article{campanella2019clinical,
  title={Clinical-grade computational pathology using weakly supervised deep learning on whole slide images},
  author={Campanella, Gabriele and Hanna, Matthew G and Geneslaw, Luke and Miraflor, Allen and Werneck Krauss Silva, Vitor and Busam, Klaus J and Brogi, Edi and Reuter, Victor E and Klimstra, David S and Fuchs, Thomas J},
  journal={Nature Medicine},
  volume={25},
  number={8},
  pages={1301--1309},
  year={2019},
  publisher={Nature Publishing Group US New York}
}

@inproceedings{he2020momentum,
  title={Momentum contrast for unsupervised visual representation learning},
  author={He, Kaiming and Fan, Haoqi and Wu, Yuxin and Xie, Saining and Girshick, Ross},
  booktitle={Proceedings of the IEEE/CVF Computer Vision and Pattern Recognition},
  pages={9729--9738},
  year={2020}
}

@article{medmnistv2,
    title={{MedMNIST} v2-A large-scale lightweight benchmark for {2D and 3D} biomedical image classification},
    author={Yang, Jiancheng and Shi, Rui and Wei, Donglai and Liu, Zequan and Zhao, Lin and Ke, Bilian and Pfister, Hanspeter and Ni, Bingbing},
    journal={Scientific Data},
    volume={10},
    number={1},
    pages={41},
    year={2023},
    publisher={Nature Publishing Group UK London}
}

@article{tschandl2018ham10000,
  title={The {HAM10000} dataset, a large collection of multi-source dermatoscopic images of common pigmented skin lesions},
  author={Tschandl, Philipp and Rosendahl, Cliff and Kittler, Harald},
  journal={Scientific Data},
  volume={5},
  pages={180161},
  year={2018},
  publisher={Nature Publishing Group}
}

@inproceedings{chen2021mocov3,
  title={An empirical study of training self-supervised {Vision Transformers}},
  author={Chen, Xinlei and Xie, Saining and He, Kaiming},
  booktitle={Proceedings of the IEEE/CVF International Conference on Computer Vision},
  pages={9640--9649},
  year={2021}
}

@inproceedings{grill2020bootstrap,
  title={{Bootstrap Your Own Latent}: A new approach to self-supervised learning},
  author={Grill, Jean-Bastien and Strub, Florian and Altch{\'e}, Florent and Tallec, Corentin and Richemond, Pierre H and Buchatskaya, Elena and Doersch, Carl and Pires, Bernardo Avila and Guo, Zhaohan Daniel and Azar, Mohammad Gheshlaghi and others},
  booktitle={Proceedings of the Conference on Neural Information Processing Systems},
  volume={33},
  pages={21271--21284},
  year={2020}
}

@inproceedings{kirillov2023segment,
  title={Segment anything},
  author={Kirillov, Alexander and Mintun, Eric and Ravi, Nikhila and Mao, Hanzi and Rolland, Chloe and Gustafson, Laura and Xiao, Tete and Whitehead, Spencer and Berg, Alexander C and Lo, Wan-Yen and others},
  booktitle={Proceedings of the IEEE/CVF International Conference on Computer Vision},
  pages={4015--4026},
  year={2023}
}

@book{edelsbrunner2010computational,
  title={Computational Topology: An Introduction},
  author={Edelsbrunner, Herbert and Harer, John},
  year={2010},
  publisher={American Mathematical Society}
}

@article{chambers2025stable,
  title={A Stable and Theoretically Grounded {Gromov-Wasserstein} Distance for {Reeb} Graph Comparison using Persistence Images},
  author={Chambers, Erin W and Meng, Guangyu},
  journal={arXiv preprint arXiv:2507.01171},
  year={2025}
}

@inproceedings{cohen2005stability,
  title={Stability of persistence diagrams},
  author={Cohen-Steiner, David and Edelsbrunner, Herbert and Harer, John},
  booktitle={Proceedings of the 21st Annual Symposium on Computational Geometry},
  pages={263--271},
  year={2005}
}

@article{qaiser2019fast,
  title={Fast and accurate tumor segmentation of histology images using persistent homology and deep convolutional features},
  author={Qaiser, Talha and Tsang, Yee-Wah and Taniyama, Daiki and Sakamoto, Naoya and Nakane, Kazuaki and Epstein, David and Rajpoot, Nasir},
  journal={Medical Image Analysis},
  volume={55},
  pages={1--14},
  year={2019},
  publisher={Elsevier}
}

@article{lawson2019persistent,
  title={Persistent homology for the quantitative evaluation of architectural features in prostate cancer histology},
  author={Lawson, Peter and Sholl, Andrew B and Brown, J Quincy and Fasy, Brittany Terese and Wenk, Carola},
  journal={Scientific Reports},
  volume={9},
  number={1},
  pages={1--12},
  year={2019},
  publisher={Nature Publishing Group}
}

@inproceedings{peng2024phg,
  title={{PHG-Net}: Persistent homology guided medical image classification},
  author={Peng, Yaopeng and Wang, Hongxiao and Sonka, Milan and Chen, Danny Z},
  booktitle={Proceedings of the IEEE/CVF Winter Conference on Applications of Computer Vision},
  pages={7583--7592},
  year={2024}
}

@incollection{wagner2011efficient,
  title={Efficient computation of persistent homology for cubical data},
  author={Wagner, Hubert and Chen, Chao and Vu{\c{c}}ini, Erald},
  booktitle={Proceedings of the Topological Methods in Data Analysis and Visualization II: Theory, Algorithms, and Applications},
  pages={91--106},
  year={2011},
  publisher={Springer}
}

@article{elston1991pathological,
  title={Pathological prognostic factors in breast cancer. {I}. The value of histological grade in breast cancer: Experience from a large study with long-term follow-up},
  author={Elston, Christopher W and Ellis, Ian O},
  journal={Histopathology},
  volume={19},
  number={5},
  pages={403--410},
  year={1991},
  publisher={Wiley Online Library}
}

@inproceedings{selvaraju2017grad,
  title={{Grad-CAM}: Visual explanations from deep networks via gradient-based localization},
  author={Selvaraju, Ramprasaath R and Cogswell, Michael and Das, Abhishek and Vedantam, Ramakrishna and Parikh, Devi and Batra, Dhruv},
  booktitle={Proceedings of the IEEE International Conference on Computer Vision},
  pages={618--626},
  year={2017}
}

@inproceedings{caron2021emerging,
  title={Emerging properties in self-supervised {Vision Transformers}},
  author={Caron, Mathilde and Touvron, Hugo and Misra, Ishan and J{\'e}gou, Herv{\'e} and Mairal, Julien and Bojanowski, Piotr and Joulin, Armand},
  booktitle={Proceedings of the IEEE/CVF International Conference on Computer Vision},
  pages={9650--9660},
  year={2021}
}

@InProceedings{Azizi_2021_ICCV,
    author    = {Azizi, Shekoofeh and Mustafa, Basil and Ryan, Fiona and Beaver, Zachary and Freyberg, Jan and Deaton, Jonathan and Loh, Aaron and Karthikesalingam, Alan and Kornblith, Simon and Chen, Ting and Natarajan, Vivek and Norouzi, Mohammad},
    title     = {Big Self-Supervised Models Advance Medical Image Classification},
    booktitle = {Proceedings of the IEEE/CVF International Conference on Computer Vision},
    month     = {October},
    year      = {2021},
    pages     = {3478-3488}
}

@article{shurrab2022self,
  title={Self-supervised learning methods and applications in medical imaging analysis: A survey},
  author={Shurrab, Saeed and Duwairi, Rehab},
  journal={PeerJ Computer Science},
  volume={8},
  pages={e1045},
  year={2022},
  publisher={PeerJ Inc.}
}

@inproceedings{zbontar2021barlow,
  title={{Barlow Twins}: Self-supervised learning via redundancy reduction},
  author={Zbontar, Jure and Jing, Li and Misra, Ishan and LeCun, Yann and Deny, St{\'e}phane},
  booktitle={Proceedings of the International Conference on Machine Learning},
  pages={12310--12320},
  year={2021}
}

@article{singh2023topological,
  title={Topological data analysis in medical imaging: Current state of the art},
  author={Singh, Yashbir and Farrelly, Colleen M and Hathaway, Quincy A and Leiner, Tim and Jagtap, Jaidip M and Carlsson, Gunnar E and Erickson, Bradley J},
  journal={Insights into Imaging},
  volume={14},
  number={1},
  pages={58},
  year={2023},
  publisher={Springer}
}

@article{zhang2020multimodal,
  title={Multimodal intelligence: Representation learning, information fusion, and applications},
  author={Zhang, Chao and Yang, Zichao and He, Xiaodong and Deng, Li},
  journal={IEEE Journal of Selected Topics in Signal Processing},
  volume={14},
  number={3},
  pages={478--493},
  year={2020},
  publisher={IEEE}
}

@article{liang2024foundations,
  title={Foundations \& trends in multimodal machine learning: Principles, challenges, and open questions},
  author={Liang, Paul Pu and Zadeh, Amir and Morency, Louis-Philippe},
  journal={ACM Computing Surveys},
  volume={56},
  number={10},
  pages={1--42},
  year={2024},
  publisher={ACM New York, NY}
}

@article{tu2024towards,
  title={Towards generalist biomedical {AI}},
  author={Tu, Tao and Azizi, Shekoofeh and Driess, Danny and Schaekermann, Mike and Amin, Mohamed and Chang, Pi-Chuan and Carroll, Andrew and Lau, Charles and Tanno, Ryutaro and Ktena, Ira and others},
  journal={NEJM AI},
  volume={1},
  number={3},
  pages={AIoa2300138},
  year={2024},
  publisher={Massachusetts Medical Society}
}

@inproceedings{carriere2020perslay,
  title={{PersLay}: A neural network layer for persistence diagrams and new graph topological signatures},
  author={Carri{\`e}re, Mathieu and Chazal, Fr{\'e}d{\'e}ric and Ike, Yuichi and Lacombe, Th{\'e}o and Royer, Martin and Umeda, Yuhei},
  booktitle={Proceedings of the International Conference on Artificial Intelligence and Statistics},
  pages={2786--2796},
  year={2020}
}

@inproceedings{ngiam2011multimodal,
  title={Multimodal deep learning},
  author={Ngiam, Jiquan and Khosla, Aditya and Kim, Mingyu and Nam, Juhan and Lee, Honglak and Ng, Andrew Y and others},
  booktitle={Proceedings of the International Conference on Machine Learning},
  volume={11},
  pages={689--696},
  year={2011}
}

@inproceedings{gao2016compact,
  title={Compact bilinear pooling},
  author={Gao, Yang and Beijbom, Oscar and Zhang, Ning and Darrell, Trevor},
  booktitle={Proceedings of the IEEE Conference on Computer Vision and Pattern Recognition},
  pages={317--326},
  year={2016}
}

@article{vaswani2017attention,
  title={Attention is all you need},
  author={Vaswani, Ashish and Shazeer, Noam and Parmar, Niki and Uszkoreit, Jakob and Jones, Llion and Gomez, Aidan N and Kaiser, {\L}ukasz and Polosukhin, Illia},
  journal={Advances in Neural Information Processing Systems},
  volume={30},
  year={2017}
}

@article{arevalo2017gated,
  title={Gated multimodal units for information fusion},
  author={Arevalo, John and Solorio, Thamar and Montes-y-G{\'o}mez, Manuel and Gonz{\'a}lez, Fabio A},
  journal={arXiv preprint arXiv:1702.01992},
  year={2017}
}

@article{lu2019vilbert,
  title={{ViLBERT}: Pretraining task-agnostic visiolinguistic representations for vision-and-language tasks},
  author={Lu, Jiasen and Batra, Dhruv and Parikh, Devi and Lee, Stefan},
  journal={Advances in Neural Information Processing Systems},
  volume={32},
  year={2019}
}

@article{shazeer2017outrageously,
  title={Outrageously large neural networks: The sparsely-gated mixture-of-experts layer},
  author={Shazeer, Noam and Mirhoseini, Azalia and Maziarz, Krzysztof and Davis, Andy and Le, Quoc and Hinton, Geoffrey and Dean, Jeff},
  journal={arXiv preprint arXiv:1701.06538},
  year={2017}
}

@article{fedus2022switch,
  title={Switch {Transformers}: Scaling to trillion parameter models with simple and efficient sparsity},
  author={Fedus, William and Zoph, Barret and Shazeer, Noam},
  journal={Journal of Machine Learning Research},
  volume={23},
  number={120},
  pages={1--39},
  year={2022}
}

@article{riquelme2021scaling,
  title={Scaling vision with sparse mixture of experts},
  author={Riquelme, Carlos and Puigcerver, Joan and Mustafa, Basil and Neumann, Maxim and Jenatton, Rodolphe and Susano Pinto, Andr{\'e} and Keysers, Daniel and Houlsby, Neil},
  journal={Advances in Neural Information Processing Systems},
  volume={34},
  pages={8583--8595},
  year={2021}
}

@article{clough2020topological,
  title={A topological loss function for deep-learning based image segmentation using persistent homology},
  author={Clough, James R and Byrne, Nicholas and Oksuz, Ilkay and Zimmer, Veronika A and Schnabel, Julia A and King, Andrew P},
  journal={IEEE Transactions on Pattern Analysis and Machine Intelligence},
  volume={44},
  number={12},
  pages={8766--8778},
  year={2020},
  publisher={IEEE}
}

@inproceedings{qi2017pointnet,
  title={{PointNet}: Deep learning on point sets for {3D} classification and segmentation},
  author={Qi, Charles R and Su, Hao and Mo, Kaichun and Guibas, Leonidas J},
  booktitle={Proceedings of the IEEE Conference on Computer Vision and Pattern Recognition},
  pages={652--660},
  year={2017}
}

@incollection{gudhi:CubicalComplex
, author    = {Pawel Dlotko}
, title     = {Cubical Complex}
, publisher = {GUDHI Editorial Board}
, edition   = {3.11.0}
, booktitle = {GUDHI User and Reference Manual}
, url       = {https://gudhi.inria.fr/doc/3.11.0/group__cubical__complex.html}
, year      = {2025}
}

@ARTICLE{11202494,
  author={Liang, Peixian and Ding, Yifan and Zhang, Yizhe and Chen, Jianxu and Zheng, Hao and Wang, Hongxiao and Zhang, Yejia and Meng, Guangyu and Weninger, Tim and Niemier, Michael and Sharon Hu, X. and Chen, Danny Z},
  journal={IEEE Transactions on Medical Imaging}, 
  title={Cell Instance Segmentation: The Devil Is in the Boundaries}, 
  year={2025},
  volume={},
  number={},
  pages={1-1},
  doi={10.1109/TMI.2025.3621093}}

@inproceedings{gu2026integrating,
  title={Integrating Multi-scale and Multi-filtration Topological Features for Medical Image Classification},
  author={Gu, Pengfei and Li, Huimin and Tang, Haoteng and Xu, Dongkuan and Enriquez, Erik and Kim, DongChul and Fu, Bin and Chen, Danny Z},
  booktitle={Proceedings of the IEEE/CVF Winter Conference on Applications of Computer Vision},
  pages={8660--8669},
  year={2026}
}

@inproceedings{gu2025self,
  title={Self pre-training with topology-and spatiality-aware masked autoencoders for {3D} medical image segmentation},
  author={Gu, Pengfei and Li, Huimin and Zhang, Yejia and Wang, Chaoli and Chen, Danny Z},
  booktitle={Proceedings of the IEEE International Conference on Bioinformatics and Biomedicine},
  pages={3608--3613},
  year={2025}
}

@inproceedings{gu2025topoimages,
  title={{TopoImages}: Incorporating local topology encoding into deep learning models for medical image classification},
  author={Gu, Pengfei and Wang, Hongxiao and Zhang, Yejia and Li, Huimin and Wang, Chaoli and Chen, Danny Z},
  booktitle={Proceedings of the 33rd ACM International Conference on Multimedia},
  pages={1938--1947},
  year={2025}
}

@inproceedings{adame2025topo,
  title={{Topo-VM-UNetV2}: Encoding Topology Into Vision {Mamba UNet} for Polyp Segmentation},
  author={Adame, Diego and Nunez, Jose A and Vazquez, Fabian and Gurrola, Nayeli and Li, Huimin and Tang, Haoteng and Fu, Bin and Gu, Pengfei},
  booktitle={Proceedings of the IEEE 38th International Symposium on Computer-Based Medical Systems},
  pages={258--263},
  year={2025}
}
}
% \input{sec/6_suppl}
% {
%     \small
%     \bibliographystyle{ieeenat_fullname}
%     \bibliography{main}
% }


\end{document}